\documentclass[11pt,a4paper]{article}
\usepackage[hyperref]{acl2019}
\usepackage{times}
\usepackage{latexsym}
\usepackage{graphicx}
\usepackage{url}
\usepackage{float}
\usepackage{multirow}
\usepackage[normalem]{ulem}

\aclfinalcopy 

\title{Beheshti-NER: Persian named entity recognition Using BERT}

\author{Ehsan Taher\thanks{*Equal contribution.} \\
	NLP Research Laboratory\\
  Shahid Beheshti University \\
Tehran, Iran \\
  \texttt{e.taher@mail.sbu.ac.ir} \\\And
  Seyed Abbas Hoseini\footnotemark[1] \\
NLP Research Laboratory\\
  Shahid Beheshti University \\
Tehran, Iran \\
  \texttt{seyeda.hoseini@mail.sbu.ac.ir} \\\AND
  Mehrnoush	Shamsfard \\
  NLP Research Laboratory\\
  Shahid Beheshti University \\
Tehran, Iran \\
  \texttt{m-shams@sbu.ac.ir} }

\begin{document}
\maketitle
\begin{abstract}
Named entity recognition is a natural language processing task to recognize and extract spans of text associated with named entities and classify them in semantic Categories.

Google BERT is a deep bidirectional language model, pre-trained on large corpora that can be fine-tuned to solve many NLP tasks such as question answering, named entity recognition, part of speech tagging and etc. In this paper, we use the pre-trained deep bidirectional network, BERT, to make a model for named entity recognition in Persian.

We also compare the results of our model with the previous state of the art results achieved on Persian NER. Our evaluation metric is CONLL 2003 score in two levels of word and phrase. This model achieved second place in NSURL-2019 task 7 competition which associated with NER for the Persian language. our results in this competition are 83.5 and 88.4 f1 CONLL score respectively in phrase and word level evaluation.

\end{abstract}

\section{Introduction}

in this paper we trained our model which is participated in NSURL-2019 task 7 competition \cite{tasknerfarsi} which associated with NER for the Persian language.

Named Entity Recognition (NER) is one of the important and basic tasks in natural language processing, assigning different parts of a text to suitable named entity categories.

There are several sets of named entity (NE) categories introduced and used in different NE tagged corpora as their tagsets. For example, Peyma's \cite{shahshahani_peyma:_2018} tagset consists of person, organization, location, date, money, percent, and time, while the Arman tagset \cite{poostchi_bilstm-crf_nodate} contains person, organization, location, facility, product, and event.

NER is one of the key parts of many downstream applications in NLP, such as question answering  \cite{Aliod2006NamedER}, information retrieval  \cite{Guo:2009:NER:1571941.1571989}, and machine translation \cite{Babych2003ImprovingMT}. As a result, the performance of NER can affect the quality of a variety of downstream applications. Furthermore, this effect is more obvious in low-resource languages because in these languages due to lack of data and tagged corpora, usually applications are implemented in pipe-line architecture unlike other languages like English which prefer to use End-to-End solutions. 

Preparing basic tools in under-resourced languages by high performance can be a good solution to such languages while we counter with lack of data issue for training such tools. 

We have trained conditional random field on the top of pre-trained bidirectional transformer BERT. Delvin et al. \cite{devlin_bert:_2019} introduced BERT as a pre-trained Bidirectional Transformer model for language understanding tasks. BERT achieved state of the art results in many tasks like question answering, language inference, and Named entity recognition.\cite{devlin_bert:_2019}

The need for large tagged data is the main problem with the recent supervised methods such as deep learning. Transfer learning can help this problem for under-resourced languages. Word embeddings approach \cite{mikolov_efficient_2013},\cite{bojanowski_enriching_2016}, \cite{joulin2017bag} and \cite{peters_elmo:_2018} are the first kind of transfer learning solutions. We use word embeddings for supervised tasks after we trained them unsupervised on large raw corpora of texts. By this, they can reduce the need for huge labeled data. They defer by BERT usually in many aspects like the fine-tuning step. After pretraining BERT on large row corpora, we fine-tune it for our specific supervised problem. While BERT tokenizes text by itself, it extracts contextualized embeddings for each token. BERT is pre-trained on 104 languages like Persian, and this is one of the big advantages of this model.  
Vaswani et al. \cite{vaswani_attention_2017} introduced transformer architecture and self-attention as an alternative for encoder-decoder recurrent neural networks architecture which could achieve state of the art results in English to German and English to France machine translation problem. Furthermore, the speed for training transformers is much less than recurrent neural networks in encoder-decoder architecture. 
CRF as a probabilistic model like hidden Markov model makes it possible to extract and consider structural dependencies among tags in data. While Encoders like BERT try to maximize likelihood by selecting best hidden representation while CRF maximizes likelihood by selecting best output tags.
We achieved 88.4\% CONLL F1 score in word-level and 83.5\% CONLL F1 score in phrase-level evaluation on Peyma dataset. We won second place in NSURL-2019 task 7 \cite{tasknerfarsi} competition for NER task.

In section 2, we talk about previous methods for NER and solutions like transfer learning to deal with under-resourced languages. Section 3 describes BERT. Section 4 explains our model in more details, discussing the training and test phases. In section 5, we show the achieved results on  experiments like evaluating our model on different datasets. Section 6 concludes the paper.

\section{Related Work}

This paper describes a deep learning method based on word embedding and transfer learning, for named entity recognition in Persian language. Thus, in this section we first discuss some related work on Persian NER, then some recent work on English NER, and then talk about some word embedding models which can be used in NER tasks via transfer learning.  

\begin{figure}[bhp]

\includegraphics[width=\linewidth]{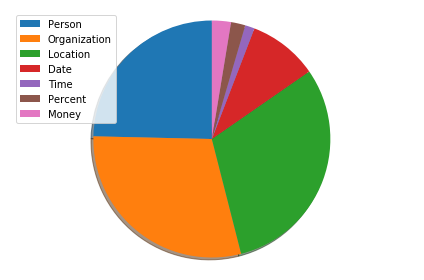}
	
	\caption{distribution of different classes in Peyma dataset}
	\label{fig:pie-peyma}
\end{figure}

\begin{figure}[bhp]

\includegraphics[width=\linewidth]{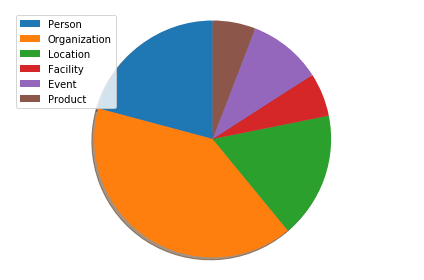}
	
	\caption{distribution of different classes in Arman dataset}
	\label{fig:pie-arman}
\end{figure}

Mortazavi and Shamsfard \cite{CSICC15_105} used a rule-based system to extract named entities for Persian languages. It was one of the first implementations for NER in Persian while no datasets existed in that time for evaluation. Poostchi et al. \cite{poostchi_bilstm-crf_nodate} introduced new annotated Persian named-entity recognition dataset named Arman. Arman contains 250,015 tokens and 7,682 sentences. Set of entity categories consists of person, organization (like banks and ministries), location, facility, product, and event. They also trained conditional random field with bidirectional LSTM on this dataset as a base-line. Shahshahani et al. \cite{shahshahani_peyma:_2018} introduced new annotated Persian named-entity recognition dataset called Peyma. Peyma contains 7,145 sentences, 302,530 tokens and 41,148 tokens with entity tags collected from 709 documents. Class distribution for both Peyma and Arman datasets are presented respectively in Fig.\ref{fig:pie-peyma} and Fig.\ref{fig:pie-arman}.

Bokaei and Mahmoudi \cite{bokaei_improved_2018} trained recurrent and convolutional neural networks with CRF on Arman dataset.  

Baevski et al. \cite{baevski_cloze-driven_2019} used a novel method for training bidirectional transformer which could over perform previous work and achieved state of the art result in English NER. 

Akbik et al.\cite{akbik_contextual_2018} used contextualized word embeddings extracted from character-level language model to solve the NER problem.  

Delvin et al. \cite{devlin_bert:_2019} introduced BERT as a pre-trained bidirectional transformer. They used and evaluated BERT on many tasks, including NER. 

Using unsupervised methods can be a promising way because the most important issue for low resource languages is the lack of labeled data while but the access to a large amount of raw texts is more probable and feasible. Today, word embeddings such as Glove \cite{pennington_glove:_2014}, word2vec \cite{mikolov_efficient_2013} , and fastText \cite{joulin2017bag} are essential parts of many methods in NLP. These models give continuous representations in n-dimensional space for each word which contain semantic information and features about that word. 

Elmo \cite{peters_elmo:_2018} introduced deep contextualized word embedding by considering the context of words. Which means words have different embedding in different contexts. Delvin et al. \cite{devlin_bert:_2019} and Radford et al.\cite{radford_language_nodate} proposed a new method with transfer self-attention blocks without the need to change in architecture for a specific problem. They suggest fine-tuning pre-trained bidirectional transformers for specific problems.  

Radford et al. \cite{radford_language_nodate} introduced a new language model called GPT.2, which could reach 55\% F1 score on the CoQA dataset without any labeled data. This approach tries to remove the need for labeled data and gives a general model to solve problems against BERT, which tries just to give a general model.

best performing models before us for NER in Persian are LSTM based models which usually come with CRF and pre-trained non-contextualized embedding layers. these models are evaluated on two common datasets for NER: PEYMA and ARMAN. Bokaei and Mahmoudi \cite{bokaei_improved_2018} and Shahshahani et al. \cite{shahshahani_peyma:_2018} had reported the best results which you can see in Table \ref{tab:compare}

\section{BERT}
\begin{figure}[t]
	\centerline{\includegraphics[width=\linewidth]{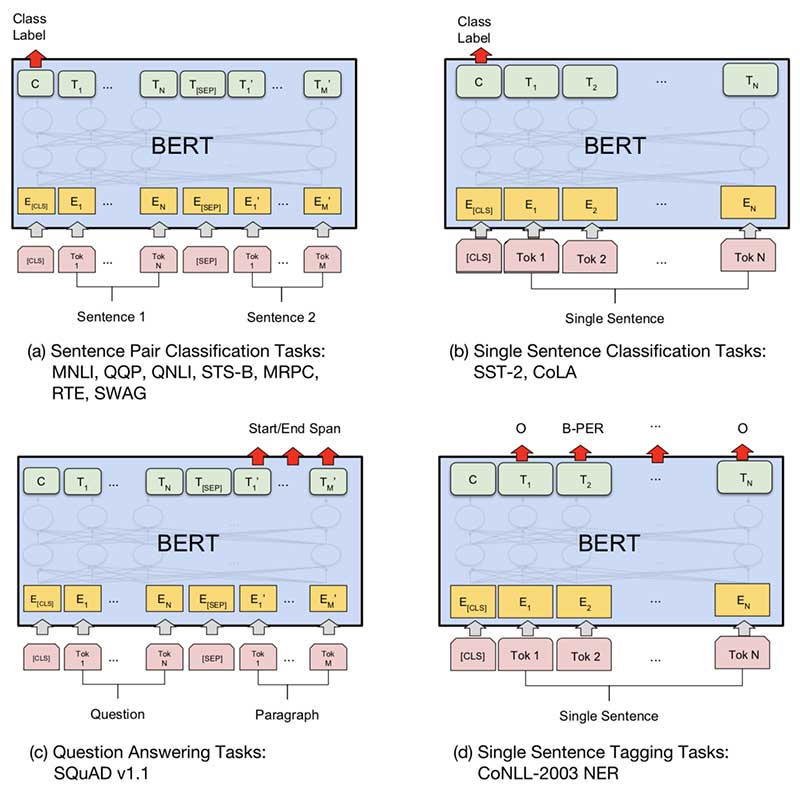}}
	\caption{fine-tuning BERT in different tasks \citep{devlin_bert:_2019}}
	\label{fig:BERT-tasks}
\end{figure}

BERT (Bidirectional Encoder Representations from Transformer) is a language model representation based on self-attention blocks. BERT is pre-trained in different language model tasks on raw unlabeled texts. The pre-trained deep bidirectional model with one output layer can reach state-of-the-art results in many tasks such as question answering and Multi-Genre Natural Language Inference. The idea is to have a general architecture which fits many problems and a pre-trained model which minimize the need for labeled data. For example, in Fig. \ref{fig:BERT-tasks} You see how BERT can be used in different tasks like question answering, sentences pair classification, single sentence classification, and single sentence tagging task. While each task has a different format of inputs and outputs. As mentioned before, one of the big advantages of BERT is that it was trained in 104 languages and Persian is one of those. Which motivate us to use it for NER in Persian.

\section{Our Proposed Model}

In this paper, we propose a method for Persian NER. In this method, we use BERT pre-trained model. As in NER task, we need to assign the most suitable tag to each token, and suitable tokenization is an important step. 

\begin{figure}[httpb]
	\centerline{\includegraphics[width=\linewidth]{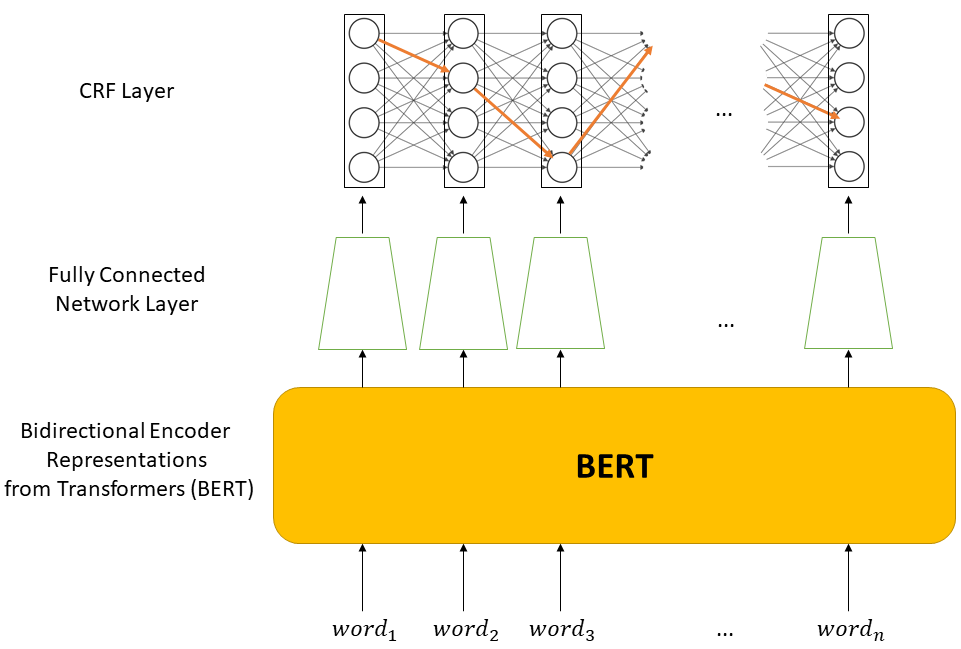}}
	
	\caption{architecture of our trained model}
	\label{fig:our-model}
\end{figure}

While BERT has its tokenization with Byte-Pair encoding and it will assign tags to its extracted tokens, we should take care of this issue. BERT extracted tokens are always equal to or smaller than our main tokens (that taken from the Step-1 \citealp{shamsfard-etal-2010-step}) because BERT takes tokens of our dataset one by one. As a result, we will have intra-tokens which take X tag (meaning don’t mention). We trained a conditional random field and fully connected layer after output representation of tokens extracted by BERT. It Is a fine-tuning step to make the entire model ready for NER task. You can see a schema of the model in Fig.\ref{fig:our-model}
. 

\section{Experiments}

We have trained and tested our model on two different datasets: Peyma \cite{shahshahani_peyma:_2018} and Arman \cite{poostchi_bilstm-crf_nodate}.We split PEYMA dataset into 5 equal subsets (Peyma contains 7145 sentences thus each subset contains 1429 sentences) and use 5-fold cross-validation. we repeated training phase 5 times separately, Each time, one of the 5 subsets is used as the test set and the remaining 4 subsets are put together to form a training set.
In all experiments, CONLL F1 score is calculated in two levels: word and phrase as a metric for evaluating the performance of model. Results of our model on Peyma and Arman datasets are given respectively in Table  \ref{tab:peyma} And Table \ref{tab:arman}.

\begin{table*}[!t]
		\resizebox{\linewidth}{!}{%
	\begin{tabular}{c|c|c|c|c|c|c|c|c|c|c|c|c|c|c|c}
		\cline{2-15}
		& \multicolumn{2}{c|}{Date}  & \multicolumn{2}{c|}{Location} & \multicolumn{2}{c|}{Money} & \multicolumn{2}{c|}{Organization} & \multicolumn{2}{c|}{Percent} & \multicolumn{2}{c|}{Person} & \multicolumn{2}{c|}{Time}\\ \cline{2-16} 
		& B-           & I-          & B-            & I-            & B-           & I-          & B-              & I-              & B-            & I-           & B-           & I-           & B-           & I-          & \multicolumn{1}{c|}{all classes} \\ \hline
		\multicolumn{1}{|c|}{word-f1}   & 84.83        & 88.44       & 91.60         & 82.39         & 95.78        & 97.59       & 89.07           & 90.29           & 94.97         & 97.13        & 93.17        & 94.25        & 83.50        & 86.48       & \multicolumn{1}{c|}{90.59}             \\ \hline
		\multicolumn{1}{|c|}{phrase-f1} & \multicolumn{2}{c|}{80.33} & \multicolumn{2}{c|}{89.75}    & \multicolumn{2}{c|}{92.54} & \multicolumn{2}{c|}{84.80}        & \multicolumn{2}{c|}{93.57}   & \multicolumn{2}{c|}{90.69}  & \multicolumn{2}{c|}{73.78} & \multicolumn{1}{c|}{87.62}     \\ \hline
	\end{tabular}}
	\caption{results of our model on Peyma dataset. Two kinds of evaluation is used, namely word and phrase level. in word level evaluation B- assigns to first token of phrase and I- is for middle and last tokens.}
	\label{tab:peyma}
\end{table*}

\begin{table*}[!t]
		\resizebox{\linewidth}{!}{%
	\begin{tabular}{c|c|c|c|c|c|c|c|c|c|c|c|c|c}
		\cline{2-13}
		& \multicolumn{2}{c|}{Event} & \multicolumn{2}{c|}{Faculty} & \multicolumn{2}{c|}{Location} & \multicolumn{2}{c|}{Organization} & \multicolumn{2}{c|}{Person} & \multicolumn{2}{c|}{Product} &                                  \\ \cline{2-14} 
		& B-           & I-          & B-            & I-           & B-            & I-            & B-              & I-              & B-           & I-           & B-            & I-           & \multicolumn{1}{c|}{all classes} \\ \hline
		\multicolumn{1}{|c|}{word-f1}   & 72.39        & 78.58       & 76.49         & 78.77        & 82.53         & 78.96         & 81.12           & 87.51           & 92.81        & 94.83        & 68.56         & 71.34        & \multicolumn{1}{c|}{84.03}       \\ \hline
		\multicolumn{1}{|c|}{phrase-f1} & \multicolumn{2}{c|}{58.45} & \multicolumn{2}{c|}{69.53}   & \multicolumn{2}{c|}{80.73}    & \multicolumn{2}{c|}{78.01}        & \multicolumn{2}{c|}{91.46}  & \multicolumn{2}{c|}{62.97}   & \multicolumn{1}{c|}{79.93}       \\ \hline
	\end{tabular}
}
	\caption{results of our model on Arman dataset.}
	\label{tab:arman}
\end{table*}

On Peyma dataset We can reach 90.59\% CONLL F1 score in phrase-level and 87.62\% F1 score in word level. Best results are seen for Percent class and worst for Time. 

On Arman dataset, We reached 79.93\% CONLL F1 score in phrase-level and 84.03\% F1 score on word-level. Best results are seen for Person class and worst for Event

One of the causes for achieving different results in each class is the amount of named entities in the datasets. As can be seen in Fig.\ref{fig:pie-peyma} and Fig.\ref{fig:pie-arman}, the number of phrases for Time and Event classes are much lower than others. 

As you see in Table  \ref{tab:compare} in both word and phrase levels, our model outperform other NER approaches for the Persian language. Unfortunately previous works  reported their results just on one of the datasets. Shahshahani et al.\cite{shahshahani_peyma:_2018} reported their results just in word level evaluation on Peyma dataset.Table \ref{tab:compare} shows that our results are 10 percent
better than Shahshahani  and colleagues on the same platform.
On the other hand Bokaei and Mahmoudi  \cite{bokaei_improved_2018} reported their
results on Arman dataset Which is lower than ours in
both word and phrase levels according to Table \ref{tab:compare}. 

\begin{table}[H]
		\resizebox{\linewidth}{!}{%
	\begin{tabular}{|c|c|c|c|c|}
		\hline
		& \multicolumn{2}{c|}{Arman} & \multicolumn{2}{c|}{Peyma} \\ \hline
		& word        & phrase       & word        & phrase       \\ \hline
		Bokaei and Mahmoudi  \cite{bokaei_improved_2018} & 81.50       & 76.79        & -           & -            \\ \hline
		Shahshahani et al.\cite{shahshahani_peyma:_2018}        & -           & -            & 80.0        & -            \\ \hline
		Beheshti-NER (Our Model)                             & \uline{\bf 84.03}       & \uline{\bf 79.93}        & \uline{\bf 90.59}       & \uline{\bf 87.62}        \\ \hline
	\end{tabular}}
	\caption{comparing results of our trained model with others}
	\label{tab:compare}
\end{table}

The results of NSURL task-7 competition is reported in two levels of evaluation, namely word and phrase levels for two subtasks: A) NER for 3- classes (Person, Location, Organization) and B) NER for all classes. for the competition, we have trained our model on PEYMA corpus in addition to another corpus which was prepared by Iran Telecommunication Research Center (ITRC). The organizers also used two kinds of in-domain and out-domain test data. Our model won second place in all of these evaluation types. 

Tables \ref{tab:aph}, \ref{tab:awo}, \ref{tab:B-ph}, \ref{tab:B-wo} and \ref{tab:B-ph-det} show the results of evaluation reported by competition for all teams which participated in the challenge. Our method is mentioned
as Beheshti-NER-1\footnote{Code is available at \url{https://github.com/sEhsanTaher/Beheshti-NER}}. 
Table \ref{tab:aph} and \ref{tab:awo} show the results for subtask A. according to the tables, we reached to 84.0\%  and 87.9\% F1 score  respectively for phrase and word level evaluations.

\begin{table}[H]
	\resizebox{\linewidth}{!}{%
		\begin{tabular}{|l|l|c|c|c|c|c|c|c|c|c|}
			\hline
			\multicolumn{2}{|c|}{\multirow{3}{*}{Team}} & \multicolumn{9}{c|}{Test Data 1}                                                              \\ \cline{3-11} 
			\multicolumn{2}{|c|}{}                      & \multicolumn{3}{c|}{In Domain} & \multicolumn{3}{c|}{Out Domain} & \multicolumn{3}{c|}{Total} \\ \cline{3-11} 
			\multicolumn{2}{|c|}{}                      & P        & R        & F1       & P         & R        & F1       & P       & R       & F1     \\ \hline
			1             & MorphoBERT                  & 88.7     & 85.5     & 87.1     & 86.3      & 83.8     & 85       & 87.3    & 84.5    & 85.9   \\ \hline
			2             & Beheshti-NER-1             & 85.3     & 84.4     & 84.8     & 84.4      & 82.6     & 83.5     & 84.8    & 83.3    & 84     \\ \hline
			3             & Team-3                      & 87.4     & 77.2     & 82       & 87.4      & 73.4     & 79.8     & 87.4    & 75      & 80.7   \\ \hline
			4             & ICTRC-NLPGroup                      & 87.5     & 76       & 81.3     & 86.2      & 69.6     & 77       & 86.8    & 72.3    & 78.9   \\ \hline
			5             & UT-NLP-IR                   & 75.3     & 68.9     & 72       & 72.3      & 60.7     & 66       & 73.6    & 64.1    & 68.5   \\ \hline
			6             & SpeechTrans                 & 41.5     & 39.5     & 40.5     & 43.1      & 38.7     & 40.8     & 42.4    & 39      & 40.6   \\ \hline
			7             & Baseline                    & 32.2     & 45.8     & 37.8     & 32.8      & 39.1     & 35.7     & 32.5    & 41.9    & 36.6   \\ \hline
	\end{tabular}
}

	\caption{Phrase-level evaluation for subtask A: 3-classes}
		\label{tab:aph}
\end{table}

\begin{table}[H]
	\resizebox{\linewidth}{!}{%
		\begin{tabular}{|l|l|c|c|c|c|c|c|c|c|c|}
			\hline
			\multicolumn{2}{|c|}{\multirow{3}{*}{Team}} & \multicolumn{9}{c|}{Test Data 1}                                                              \\ \cline{3-11} 
			\multicolumn{2}{|c|}{}                      & \multicolumn{3}{c|}{In Domain} & \multicolumn{3}{c|}{Out Domain} & \multicolumn{3}{c|}{Total} \\ \cline{3-11} 
			\multicolumn{2}{|c|}{}                      & P        & R        & F1       & P         & R        & F1       & P       & R       & F1     \\ \hline
			1             & MorphoBERT                  & 92.5     & 86.7     & 89.5     & 91.5      & 84       & 87.6     & 92.1    & 85.2    & 88.5   \\ \hline
			2             & Beheshti-NER-1             & 90.5     & 87.2     & 88.8     & 89.7      & 85       & 87.3     & 90.1    & 85.8    & 87.9   \\ \hline
			3             & Team-3                      & 89.2     & 79.5     & 84.1     & 89.5      & 74.7     & 81.4     & 89.3    & 76.9    & 82.7   \\ \hline
			4             & ICTRC-NLPGroup                      & 90.1     & 78.2     & 83.7     & 88.7      & 70.2     & 78.4     & 89.4    & 73.5    & 80.7   \\ \hline
			5             & UT-NLP-IR                   & 87.3     & 71.9     & 78.9     & 86.4      & 61.1     & 71.6     & 86.9    & 65.7    & 74.8   \\ \hline
			6             & SpeechTrans                 & 66.8     & 38.3     & 48.7     & 66.2      & 35.2     & 46       & 66.6    & 36.4    & 47     \\ \hline
			7             & Baseline                    & 46.2     & 42.6     & 44.3     & 45.2      & 35.1     & 39.5     & 45.9    & 38.4    & 41.8   \\ \hline
	\end{tabular}
}

	\caption{Word-level evaluation for subtask A: 3-classes}
		\label{tab:awo}
\end{table}

results for subtask B is given in Table \ref{tab:B-ph} and \ref{tab:B-wo}. we can achieve 83.5\% and 88.4\% F1 score respectively for phrase and word level evaluation.
\begin{table}[H]
	\resizebox{\linewidth}{!}{%
		\begin{tabular}{|l|l|c|c|c|c|c|c|c|c|c|}
			\hline
			\multicolumn{2}{|c|}{\multirow{3}{*}{Team}} & \multicolumn{9}{c|}{Test Data 1}                                                              \\ \cline{3-11} 
			\multicolumn{2}{|c|}{}                      & \multicolumn{3}{c|}{In Domain} & \multicolumn{3}{c|}{Out Domain} & \multicolumn{3}{c|}{Total} \\ \cline{3-11} 
			\multicolumn{2}{|c|}{}                      & P        & R        & F1       & P         & R        & F1       & P       & R       & F1     \\ \hline
			1             & MorphoBERT                  & 88.4     & 84.8     & 86.6     & 86        & 83.1     & 84.5     & 87      & 83.8    & 85.4   \\ \hline
			2             & Beheshti-NER-1             & 84.8     & 83.6     & 84.2     & 83.9      & 82       & 83       & 84.3    & 82.7    & 83.5   \\ \hline
			3             & Team-3                      & 87.4     & 77.3     & 82       & 87.3      & 72.8     & 79.4     & 87.3    & 74.7    & 80.5   \\ \hline
			4             & ICTRC-NLPGroup                      & 87       & 76.1     & 81.2     & 86.2      & 70.2     & 77.4     & 86.5    & 72.7    & 79     \\ \hline
			5             & UT-NLP-IR                   & 77.3     & 70.2     & 73.6     & 74.1      & 61.9     & 67.5     & 75.5    & 65.4    & 70.1   \\ \hline
			6             & SpeechTrans                 & 38       & 34.5     & 36.2     & 38.9      & 33.6     & 36       & 38.5    & 34      & 36.1   \\ \hline
			7             & Baseline                    & 32.8     & 45.7     & 38.2     & 32        & 38.1     & 34.8     & 32.4    & 41.3    & 36.3   \\ \hline
	\end{tabular}
}

	\caption{Phrase-level evaluation for subtask B: 7-classes}
		\label{tab:B-ph}
\end{table}

\begin{table}[H]
	\resizebox{\linewidth}{!}{%
		\begin{tabular}{|l|l|c|c|c|c|c|c|c|c|c|}
			\hline
			\multicolumn{2}{|c|}{\multirow{3}{*}{Team}} & \multicolumn{9}{c|}{Test Data 1}                                                              \\ \cline{3-11} 
			\multicolumn{2}{|c|}{}                      & \multicolumn{3}{c|}{In Domain} & \multicolumn{3}{c|}{Out Domain} & \multicolumn{3}{c|}{Total} \\ \cline{3-11} 
			\multicolumn{2}{|c|}{}                      & P        & R        & F1       & P         & R        & F1       & P       & R       & F1     \\ \hline
			1             & MorphoBERT                  & 94       & 89.1     & 91.5     & 91.8      & 85.7     & 88.6     & 92.8    & 87.1    & 89.9   \\ \hline
			2             & Beheshti-NER-1             & 91.4     & 87.3     & 89.3     & 89.7      & 85.7     & 87.7     & 90.4    & 86.5    & 88.4   \\ \hline
			3             & Team-3                      & 91.3     & 84.1     & 87.5     & 90.9      & 77.9     & 83.9     & 91.1    & 80.7    & 85.5   \\ \hline
			4             & ICTRC-NLPGroup                      & 89.2     & 83.1     & 86.1     & 89.8      & 76.5     & 82.6     & 89.7    & 79.4    & 84.2   \\ \hline
			5             & UT-NLP-IR                   & 92.7     & 79.3     & 85.4     & 91.1      & 68.4     & 78.1     & 91.9    & 73.1    & 81.4   \\ \hline
			6             & SpeechTrans                 & 76.1     & 32.9     & 45.9     & 74.9      & 30.3     & 43.2     & 75.7    & 31.5    & 44.5   \\ \hline
			7             & Baseline                    & 50.6     & 47.8     & 49.2     & 42.6      & 35.1     & 38.5     & 46.5    & 40.9    & 43.5   \\ \hline
		\end{tabular}
	}

	\caption{Word-level evaluation for subtask B: 7-classes}
		\label{tab:B-wo}
\end{table}

details of evaluation for each class in subtask B is given in Table \ref{tab:B-wo}. as you see all teams have higher scores in Percent class and the worst score for many teams is for Time class.
\begin{table}[H]
	\resizebox{\linewidth}{!}{%
		\begin{tabular}{|l|l|c|c|c|c|c|c|c|c|}
			\hline
			\multicolumn{2}{|c|}{\multirow{2}{*}{Team}} & \multicolumn{8}{c|}{Test Data 1}                           \\ \cline{3-10} 
			\multicolumn{2}{|c|}{}                      & PER  & ORG   & LOC  & DAT  & TIM  & MON  & PCT  & Total F1 \\ \hline
			1             & MorphoBERT                  & 90.4 & 80.3  & 87.1 & 78.9 & 71   & 93.6 & 96.8 & 85.4     \\ \hline
			2             & Beheshti-NER-1             & 81.8 & 80.8  & 88   & 77.8 & 75.8 & 85.1 & 91.6 & 83.5     \\ \hline
			3             & Team-3                      & 79.9 & 77.2  & 83.9 & 74.7 & 64.3 & 92.1 & 97.4 & 80.5     \\ \hline
			4             & ICTRC-NLPGroup                      & 76.2 & 75.93 & 82.8 & 76   & 67.1 & 91.3 & 93.6 & 79       \\ \hline
			5             & UT-NLP-IR                   & 63.4 & 58.8  & 78.2 & 76.1 & 69.1 & 84.5 & 93.5 & 70.1     \\ \hline
			6             & SpeechTrans                 & 24.3 & 23.5  & 63.1 & 12   & 4.1  & 0.3  & 0.7  & 36.1     \\ \hline
			7             & Baseline                    & 23.5 & 38.1  & 44.2 & 41.6 & 30.3 & 13.7 & 36.6 & 36.3     \\ \hline
		\end{tabular}
	}

	\caption{Details of phrase-level evaluation for subtask B: 7-classes}
		\label{tab:B-ph-det}
\end{table}

\section{Conclucion}

in this work we fine-tuned the pre-trained BERT model with a CRF layer in NER task for Persian language. our trained model achieved best results compared to the previous ones and ranked as the second team in NSURL competition. this work present BERT as a good transfer learning solution for solving low resource problems. 

results show that our model could outperform previous methods with a dramatic difference. the reason for this could be using a big pre-trained model, BERT, which achieved state of the art results in English and proved to perform well with a less amount of data for training.

\bibliography{acl2019}
\bibliographystyle{acl_natbib}

\end{document}